\pdfoutput=1

\documentclass[11pt]{article}

\usepackage[final]{acl}

\usepackage{times}
\usepackage{latexsym}

\usepackage[T1]{fontenc}

\usepackage[utf8]{inputenc}

\usepackage{microtype}

\usepackage{inconsolata}

\usepackage{graphicx}
\usepackage{multirow}
\usepackage{xcolor}
\usepackage{colortbl}
\usepackage{booktabs}
\usepackage{amsmath}
\usepackage{float}
\usepackage{enumitem}

\title{Broken Words, Broken Performance:\\ Effect of Tokenization on Performance of LLMs}

\author{Sachin Pawar, Manoj Apte, Kshitij Jadhav, Girish K. Palshikar\thanks{Work done while working at TCS Research}, Nitin Ramrakhiyani \\
TCS Research, Tata Consultancy Services Limited, India. \\
\texttt{\small\{sachin7.p, manoj.apte, kshitij.jadhav, nitin.ramrakhiyani\}@tcs.com}, \texttt{\small girishpalshikar@gmail.com}}

\begin{document}
\maketitle
\begin{abstract}
Tokenization is the first step in training any Large Language Model (LLM), where the text is split into a sequence of tokens as per the model's fixed vocabulary. This tokenization in LLMs is different from the traditional tokenization in NLP where the text is split into a sequence of {\em natural} words. In LLMs, a natural word may also be broken into multiple tokens due to limited vocabulary size of the LLMs (e.g., Mistral's tokenizer splits {\it martial} into {\it mart} and {\it ial}). In this paper, we hypothesize that such breaking of natural words negatively impacts LLM performance on various NLP tasks. To quantify this effect, we propose a set of penalty functions that compute a tokenization penalty for a given text for a specific LLM, indicating how {\em bad} the tokenization is. We establish statistical significance of our hypothesis on multiple NLP tasks for a set of different LLMs.
\end{abstract}

\section{Introduction}
Recently, Large Language Models (LLMs) have been showing remarkable language understanding and generation capabilities. However, it is also observed that LLMs tend to produce inaccurate or unexpected results for some specific queries, the reasons for which can be traced back to the initial step of tokenization~\cite{wang2024tokenization, karpathy2024}. Tokenization is the very first step in training any LLM where the text is split into a sequence of tokens as per the model's fixed vocabulary. This vocabulary is generally determined based on a different training corpus (and often much smaller) than the model's actual training corpus, using techniques such as Byte-Pair Encoding (BPE)~\cite{sennrich2016neural}. Due to this fixed and limited vocabulary of tokens, LLMs often break a {\em natural} word into multiple tokens. Here, by {\em natural} word, we mean the words obtained by traditional tokenization\footnote{\url{https://spacy.io/api/tokenizer/}} in NLP. Nature of this tokenization significantly influences how the text is interpreted, processed and generated by the models. 

\begin{figure}
    \centering
    \includegraphics[width=\columnwidth,height=0.55\columnwidth]{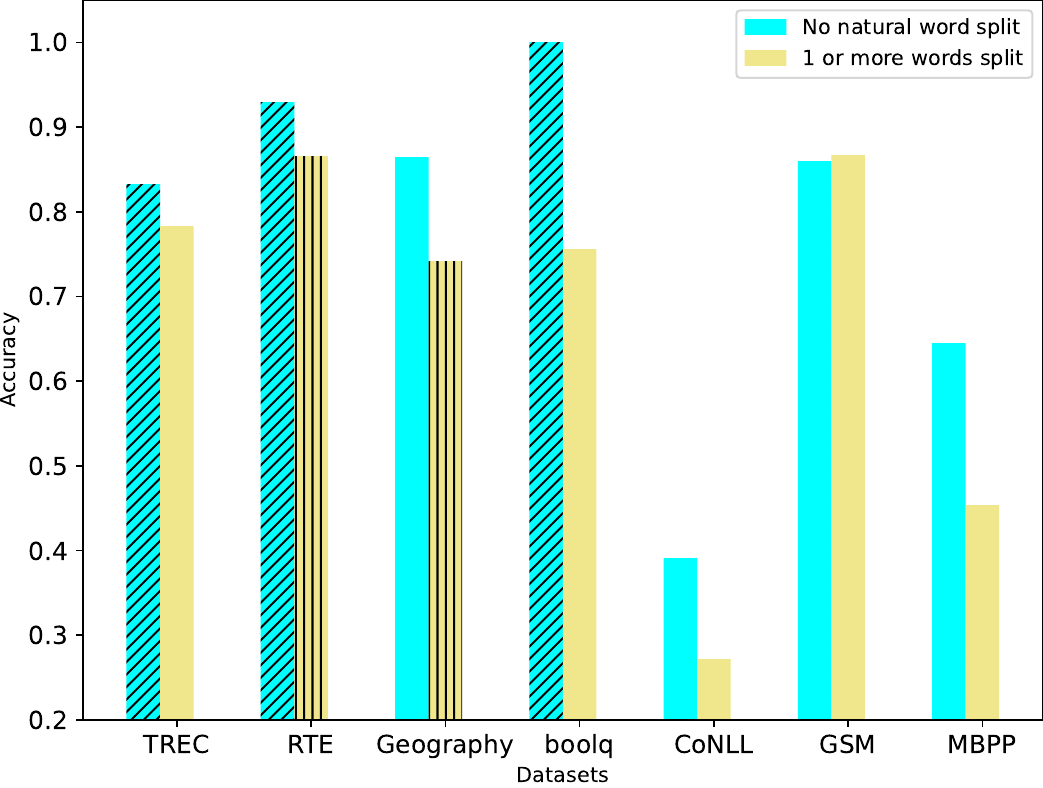}
    \caption{Effect of splitting words into multiple tokens}
    \label{figExamplePerformance}
\end{figure}

Consider the word {\it unhappiness}. A human can break this long word into subwords intuitively like {\it un}, {\it happy} and {\it ness} which might lead to preserving the morphological structure. However, the tokenizer of an LLM (say Phi-3.5-mini-instruct) may break it as {\it unh}, {\it app}, {\it iness}. 
Since the model has learned the sequence of such tokens, the model's understanding of the query may shift subtly or drastically based on the specific tokens created by the tokenizer. When such shifts accumulate across an input they might create inconsistencies and in turn measurable change in the output of the model. For a variety of NLP tasks (Section~\ref{secDatasets}), we observed that whenever there is no such breaking of any natural word in the input text, the model performance is generally better than the case when at least one word is split into multiple tokens (see Figure~\ref{figExamplePerformance}).
\begin{table}[t]\small
    \centering
    \begin{tabular}{p{0.95\columnwidth}}
    \toprule
    \textbf{P}: {\it{\color{blue}Ostriches} put their heads into the sand to avoid the wind.} \\
    \textbf{H1}: {\it{\color{blue}Ostriches} {\bf\color{blue}bury} their heads in the sand.} \\
    \textbf{H2}: {\it{\color{blue}Ostriches} {\bf hide} their heads in the sand.} \\
    \midrule
    Phi-3.5-mini-instruct predicts {\color{red}No-Entailment} for $\langle$P, H1$\rangle$ \\
    Phi-3.5-mini-instruct predicts {\color{teal}Entailment} for $\langle$P, H2$\rangle$ \\
    \midrule
    Natural words which are split into multiple tokens:\\
    \textit{\color{blue}Ostriches} $\Rightarrow$ \textit{\_O}, \textit{str}, \textit{ich}, \textit{es}~;~\textit{\color{blue}bury} $\Rightarrow$ \textit{\_b}, \textit{ury} \\
    \bottomrule
    \end{tabular}
    \caption{An example where the prediction is corrected with a small change (\textit{bury} $\Rightarrow$\textit{hide}) in the input}
    \label{tabRTEExample}
\end{table}
We also observed a few examples where a small change in the input text to avoid such breaking of natural words (e.g., by using synonyms for such words) can lead to better performance. Table~\ref{tabRTEExample} shows one such example for the textual entailment task where the LLM output is corrected by a small change in the input.

Thus, one important question arises -- how do small differences in tokenization lead to significant downstream effects on model behavior? In this paper, we attempt to answer this question by defining and quantifying ``tokenization penalty'' as the change in model performance or output quality that is attributable to tokenization induced distortions. One of the most relevant related work is by~\citet{land2024fishing} where they use token embeddings to find a set of {\em under-trained} tokens. One of our proposed penalty functions is designed based on their work. Another relevant work is by~\citet{wang2024tokenization} where they create an adversarial dataset for an LLM from the tokenization perspective. However, the constructed dataset covers only one kind of NLP task (sentence-based QA) and it contains artificially introduced errors (e.g., concatenating \textit{moves} and \textit{table} to \textit{movestable}). On the other hand, we explore multiple types of NLP tasks (Section~\ref{secDatasets}) and analyse the original input text as it is, to quantify tokenization related issues in it. 
There is another line of research~\citep{chizhov2024bpe,lian2025scaffold,lian2025lbpe} which deals with improving the basic BPE tokenization algorithm. 
This is complimentary to our work in the sense that they attempt to improve tokenization quality by eliminating noisy tokens whereas we attempt to quantify the quality of tokenization (obtained by any tokenization algorithm) for a specific text.

Our specific contributions are:
\begin{itemize}[leftmargin=*, itemsep=0pt, topsep=0pt]
    \item multiple {\em tokenization penalty} functions to quantify the effect of ``bad'' tokenization (Section~\ref{secTokenizationPenaltyFunction})
    \item statistical significance tests to measure the effect of tokenization penalty on the performance of LLMs (Section~\ref{secStatTest})
\end{itemize}




\section{Tokenization Penalty Functions}\label{secTokenizationPenaltyFunction}
We propose 4 different penalty functions where a penalty is calculated for each natural word in the LLM input text. If any natural word corresponds to a single token as per the LLM's tokenizer, there is no penalty considered. Here, we use the spaCy ~\cite{honnibal2020spacy} 
tokenizer 
to get the list of tokens in the input text and each purely alphabetic token
\footnote{Matching the regex \texttt{\^{}[A-Za-z]+}}
is considered as a valid natural word. Consider a text $T$ which is an input to LLM $M$. Let $w$ be a valid natural word in $T$ which is split into $k$ tokens $t_{w_1}\cdots t_{w_k}$. Let the tokens to the left of $t_{w_i}$ in $T$ be $T_{<w_i}$. 
Let $\vec{t}$ be the vector representation of the token $t$ as 
per $M$'s output embedding matrix (following~\citet{land2024fishing}).\\

\begin{table}[]\small
    \centering
    \begin{tabular}{p{0.95\columnwidth}}
\toprule
\textbf{S1}: \textit{{\bf Bacteria} is winning the war against {\bf antibiotics}.}\\
\textbf{S2}: \textit{It is winning the war against {\bf antibiotics}.}\\
\midrule
Natural words which are split into multiple tokens:\\
\textit{Bacteria} $\Rightarrow$ \textit{\_B}, \textit{acter}, \textit{ia}~;~
\textit{antibiotics} $\Rightarrow$ \textit{\_ant}, \textit{ib}, \textit{iot}, \textit{ics} \\
\bottomrule
    \end{tabular}
    \caption{Tokenization example (Phi-3.5-mini-instruct)}
    \label{tabTokExample}
\end{table}

\noindent\textbf{Penalty based on Token Anomaly Scores} ({\small $\boldsymbol{AS(w)}$}): The intuition behind this penalty function is -- higher the anomaly score of any token, higher is the penalty. We use Isolation Forest ($IF$)~\cite{liu2008isolation} over all the tokens in $M$'s vocabulary to compute anomaly score for each token and normalize the scores to lie in $[0,1]$ (see~\ref{secIF}). Overall penalty for the word $w$ is computed as:
\begin{equation}\small
    AS(w) = \frac{1}{k}\sum_{i=1}^k AnomalyScore_{IF}(\vec{t}_{w_i})
\end{equation}
\noindent\textbf{Penalty based on Similarity with Under-trained Tokens} ({\small $\boldsymbol{UT(w)}$}): It is motivated by~\citet{land2024fishing} that a token is {\em under-trained} if its embedding is closer to the average embeddings ($\vec{u}$) of unused tokens in the model's vocabulary. So in this case the penalty for word $w$ is:
\begin{equation}\small
    UT(w) = \frac{1}{k}\sum_{i=1}^k \left(1-CosineDistance(\vec{t}_{w_i}, \vec{u})\right)
\end{equation}
\noindent\textbf{Penalty based on Pairwise Distance between Tokens} ({\small $\boldsymbol{PD(w)}$}): The intuition behind this penalty function is -- higher the distance between two consecutive tokens of $w$, higher is the penalty.
\begin{equation}\small
    PD(w) = \frac{1}{k-1}\sum_{i=1}^{k-1} CosDist(\vec{t}_{w_i}, \vec{t}_{w_{i+1}}))
\end{equation}
\noindent\textbf{Contextual Penalty} ({\small $\boldsymbol{CP(w)}$}): All the earlier penalty functions are {\em non-contextual}, i.e., the penalty for the word $w$ is independent of its context within $T$. We propose another penalty function which is {\em contextual}, i.e., the penalty for the same word $w$ may vary if its context changes. Here, the intuition is that if the left context of $w$ in $T$ is such that the model $M$ is less {\em perplexed} to see the tokens in $w$, then the tokenization penalty should be lower, and vice versa. We quantify this penalty using conditional probability of the tokens in $w$ given their left context, as per the model $M$. E.g., in Table~\ref{tabTokExample}, the penalty for the word {\it antibiotics} would be much higher in sentence S2 as compared to S1 in which the context contains the word {\it Bacteria}. 
Also, part-of-speech (POS) tag of $w$ may change based on its context in $T$. We hypothesize that certain POS tags (verbs, common nouns, adjectives and adverbs) contribute more to the overall meaning compared to certain other POS tags like proper nouns, prepositions, etc. 
Hence, tokenization penalty for $w$ with POS tag $p(w)$ is multiplied by POS importance weight $wt_{p(w)}$\footnote{We use $wt_{p(w)}=2$ for verbs, common nouns, adjectives, adverbs and $wt_{p(w)}=1$ for other POS tags.}. 
So, the overall penalty\footnote{Though Eq.~\ref{eqCP} looks similar to {\em perplexity}, there is a key difference - unlike perplexity which considers all the tokens in a text, $CP$ considers only those tokens which are part of some split natural word. See Appendix~\ref{secPPL} for more details.} for the word $w$ is:
\begin{equation}\small\label{eqCP}
    CP(w) = \frac{1}{k}\left(wt_{p(w)}\cdot\sum_{i=1}^k \left(1-P_M(t_{w_i}|T_{<w_i})\right)\right)
\end{equation}
Here, $P_M$ denotes the next token probability as per model $M$ and $p(w)$ denotes the POS tag of $w$. 
Appendix Table~\ref{tabExamplePenaltyFunctions} shows various tokenization penalty functions computed for two example sentences.

\noindent\textbf{Aggregation Techniques}: All the above functions compute the tokenization penalty for a single word $w$ appearing in text $T$. We explore 4 different aggregation techniques to compute the tokenization penalty for $T$ aggregating over penalties for all the words in it -- 
\begin{enumerate}[label=\roman*, itemsep=0pt, topsep=0pt]
    \item \textbf{sum}: addition of all the word penalties
    \item \textbf{mean}: average of all the word penalties
    \item \textbf{max}: maximum among all the word penalties
    \item \textbf{top\_K}: average of the top $K$ word penalties
\end{enumerate}

\section{Measuring Effect of Bad Tokenization}\label{secStatTest}
To quantify the effect of ``bad'' tokenization on the performance of LLMs, we perform a statistical test. Consider an NLP task (e.g., classification, NER) and a corresponding dataset $D$ with $n$ instances. For each instance in $D$, we compute its tokenization penalty. We then use an LLM 
to generate the output for each of these instances\footnote{Appendix Table~\ref{tabLLMPrompts} shows the detailed prompts.} and prepare two sets of tokenization penalties -- (i) $C$ (tokenization penalties for instances where the LLM produced \textit{correct} output) and (ii) $I$ (tokenization penalties for instances where the LLM produced \textit{incorrect} output). We then use a one-sided two-sample \textbf{Student's \textit{t}-test} with the null and alternate hypotheses:
\begin{itemize}[itemsep=0pt, topsep=0pt]
    \item[$H_0$] $Mean(I) = Mean(C)$ (average tokenization penalties are same for both correct as well as incorrect instances)
    \item[$H_1$] $Mean(I) > Mean(C)$ (average tokenization penalties for incorrect instances are higher than the correct instances)
\end{itemize}
We also perform the \textbf{Mann–Whitney U test} in a similar manner, as it is a non-parametric method that does not assume normality.


\section{Experiments}
We evaluate the proposed tokenization penalty functions on 7 different NLP tasks and 4 different LLMs -- 
Phi~\cite{abdin2024phi3technicalreporthighly}, Mistral~\cite{jiang2023mistral7b}, Qwen~\cite{qwen2025qwen25technicalreport}, and Llama~\cite{grattafiori2024llama3herdmodels} having varying vocabulary sizes (see Figure~\ref{figLLMsVocab}). 
\begin{table*}[t]\small
    \centering
    \begin{tabular}{p{1cm}lc|c|c|cc|cc|cc|cccc}
    \toprule
    \multirow{2}{*}{\textbf{Dataset}} & \multirow{2}{*}{\textbf{Model}} & \multirow{2}{*}{\textbf{Acc}} & \multirow{2}{*}{\textbf{B1}} & \multirow{2}{*}{\textbf{B2}} & \multicolumn{2}{c|}{\textbf{AS}} & \multicolumn{2}{c|}{\textbf{UT}} & \multicolumn{2}{c|}{\textbf{PD}} & \multicolumn{4}{c}{\textbf{CP}} \\
    \cline{6-15}
     &  &  &  &  & \textbf{sum} & \textbf{max} & \textbf{sum} & \textbf{max} & \textbf{sum} & \textbf{max} & \textbf{sum} & \textbf{avg} & \textbf{max} & \textbf{top3} \\
    \midrule
    \multirow{4}{*}{TREC} & Phi & .796 & \cellcolor{green!25}.005 & .100 & \cellcolor{orange!20}.097 & \cellcolor{orange!20}.098 & .144 & .154 & \cellcolor{orange!20}.084 & \cellcolor{orange!20}.083 & \cellcolor{orange!20}.068 & .746 & \cellcolor{green!25}.040 & \cellcolor{green!25}.043 \\
                      & Mistral & .710 & .258 & .123 & \cellcolor{orange!20}.055 & \cellcolor{green!25}.007 & .120 & \cellcolor{green!25}.033 & .136 & \cellcolor{green!25}.037 & .110 & \cellcolor{green!25}.040 & \cellcolor{green!25}.033 & \cellcolor{green!25}.028 \\
                      & Qwen & .776 & .182 & \cellcolor{orange!20}.096 & .112 & \cellcolor{orange!20}.091 & \cellcolor{orange!20}.098 & \cellcolor{orange!20}.083 & \cellcolor{orange!20}.097 & \cellcolor{orange!20}.077 & \cellcolor{green!25}.037 & \cellcolor{green!25}.019 & \cellcolor{green!25}.039 & \cellcolor{green!25}.033 \\
                      & Llama & .754 & .276 & .184 & .162 & .133 & .198 & .173 & .177 & .143 & \cellcolor{orange!20}.079 & \cellcolor{green!25}.026 & \cellcolor{orange!20}.076 & \cellcolor{orange!20}.069 \\
    \midrule
    \multirow{4}{*}{RTE} & Phi & .865 & .490 & .225 & .264 & .569 & .217 & .147 & .237 & .250 & .109 & .150 & .263 & .242 \\
                      & Mistral & .818 & .933 & .987 & .989 & .733 & .982 & .395 & .987 & .514 & .830 & .632 & .560 & .459 \\
                      & Qwen & .895 & .871 & .933 & .924 & .528 & .949 & .586 & .938 & .576 & .781 & .542 & .670 & .574 \\
                      & Llama & .748 & .999 & .999 & .999 & .974 & .999 & .968 & .999 & .966 & .971 & .291 & .639 & .469 \\
    \midrule
    \multirow{4}{*}{boolq} & Phi & .757 & .155 & .178 & .182 & \cellcolor{green!25}.009 & .260 & .810 & .174 & \cellcolor{orange!20}.089 & \cellcolor{green!25}.011 & \cellcolor{green!25}.001 & \cellcolor{green!25}.000 & \cellcolor{green!25}.000 \\
                      & Mistral & .755 & .271 & .126 & \cellcolor{orange!20}.098 & \cellcolor{green!25}.009 & .148 & .262 & .116 & \cellcolor{orange!20}.074 & \cellcolor{green!25}.045 & \cellcolor{green!25}.006 & \cellcolor{green!25}.015 & \cellcolor{green!25}.012 \\
                      & Qwen & .773 & .571 & .601 & .604 & .656 & .621 & .827 & .596 & .737 & \cellcolor{orange!20}.059 & \cellcolor{green!25}.037 & \cellcolor{green!25}.009 & \cellcolor{green!25}.018 \\
                      & Llama & .798 & .206 & .370 & .310 & .456 & .433 & .844 & .363 & .671 & .107 & .122 & \cellcolor{green!25}.018 & \cellcolor{green!25}.012 \\
    \midrule
    \multirow{4}{*}{\footnotesize Geography} & Phi & .751 & \cellcolor{orange!20}.082 & \cellcolor{green!25}.015 & \cellcolor{green!25}.026 & \cellcolor{green!25}.000 & .136 & \cellcolor{green!25}.007 & \cellcolor{green!25}.024 & \cellcolor{green!25}.000 & \cellcolor{green!25}.000 & \cellcolor{green!25}.000 & \cellcolor{green!25}.000 & \cellcolor{green!25}.000 \\
                      & Mistral & .790 & .258 & \cellcolor{green!25}.000 & \cellcolor{green!25}.000 & \cellcolor{green!25}.001 & \cellcolor{green!25}.013 & \cellcolor{green!25}.017 & \cellcolor{green!25}.000 & \cellcolor{green!25}.000 & \cellcolor{green!25}.000 & \cellcolor{green!25}.000 & \cellcolor{green!25}.000 & \cellcolor{green!25}.000 \\
                      & Qwen & .726 & .342 & \cellcolor{green!25}.001 & \cellcolor{green!25}.001 & \cellcolor{green!25}.000 & \cellcolor{green!25}.003 & \cellcolor{green!25}.000 & \cellcolor{green!25}.001 & \cellcolor{green!25}.000 & \cellcolor{green!25}.000 & \cellcolor{green!25}.000 & \cellcolor{green!25}.000 & \cellcolor{green!25}.000 \\
                      & Llama & .699 & .554 & .998 & .998 & .888 & .997 & .840 & .997 & .775 & .756 & .645 & .207 & \cellcolor{orange!20}.078 \\
    \midrule
    \multirow{4}{*}{CoNLL} & Phi & .280 & \cellcolor{green!25}.000 & \cellcolor{green!25}.000 & \cellcolor{green!25}.000 & \cellcolor{green!25}.001 & \cellcolor{green!25}.000 & \cellcolor{green!25}.000 & \cellcolor{green!25}.000 & \cellcolor{green!25}.005 & \cellcolor{green!25}.000 & .336 & \cellcolor{green!25}.001 & \cellcolor{green!25}.012 \\
                      & Mistral & .336 & \cellcolor{green!25}.000 & \cellcolor{green!25}.000 & \cellcolor{green!25}.000 & \cellcolor{green!25}.000 & \cellcolor{green!25}.000 & \cellcolor{green!25}.000 & \cellcolor{green!25}.000 & \cellcolor{green!25}.000 & \cellcolor{green!25}.000 & .358 & \cellcolor{green!25}.000 & \cellcolor{green!25}.000 \\
                      & Qwen & .481 & \cellcolor{green!25}.000 & \cellcolor{green!25}.000 & \cellcolor{green!25}.000 & \cellcolor{green!25}.000 & \cellcolor{green!25}.000 & \cellcolor{green!25}.000 & \cellcolor{green!25}.000 & \cellcolor{green!25}.000 & \cellcolor{green!25}.000 & .114 & \cellcolor{green!25}.000 & \cellcolor{green!25}.030 \\
                      & Llama & .389 & \cellcolor{green!25}.000 & \cellcolor{green!25}.000 & \cellcolor{green!25}.000 & \cellcolor{green!25}.000 & \cellcolor{green!25}.000 & \cellcolor{green!25}.000 & \cellcolor{green!25}.000 & \cellcolor{green!25}.000 & \cellcolor{green!25}.000 & .312 & \cellcolor{green!25}.000 & \cellcolor{green!25}.011 \\
    \midrule
    \multirow{4}{*}{GSM} & Phi & .867 & \cellcolor{green!25}.000 & \cellcolor{orange!20}.069 & \cellcolor{orange!20}.092 & .464 & \cellcolor{orange!20}.075 & .265 & \cellcolor{orange!20}.061 & .457 & \cellcolor{green!25}.009 & .666 & .205 & \cellcolor{green!25}.040 \\
                      & Mistral & .640 & \cellcolor{green!25}.000 & \cellcolor{green!25}.047 & \cellcolor{orange!20}.084 & .512 & \cellcolor{orange!20}.065 & .382 & \cellcolor{green!25}.044 & .303 & .119 & .998 & .434 & .265 \\
                      & Qwen & .916 & \cellcolor{green!25}.000 & .690 & .753 & .509 & .694 & .374 & .686 & .387 & .426 & .929 & .493 & .230 \\
                      & Llama & .813 & \cellcolor{green!25}.000 & .156 & .176 & .172 & .177 & .214 & .154 & .154 & .134 & .954 & .198 & .116 \\
    \midrule
    \multirow{4}{*}{MBPP} & Phi & .553 & .131 & \cellcolor{green!25}.000 & \cellcolor{green!25}.000 & \cellcolor{green!25}.000 & \cellcolor{green!25}.000 & \cellcolor{green!25}.000 & \cellcolor{green!25}.000 & \cellcolor{green!25}.000 & \cellcolor{green!25}.000 & \cellcolor{green!25}.000 & \cellcolor{green!25}.000 & \cellcolor{green!25}.002 \\
                      & Mistral & .372 & \cellcolor{green!25}.003 & \cellcolor{green!25}.000 & \cellcolor{green!25}.000 & \cellcolor{green!25}.001 & \cellcolor{green!25}.000 & \cellcolor{green!25}.000 & \cellcolor{green!25}.000 & \cellcolor{green!25}.001 & \cellcolor{green!25}.004 & \cellcolor{green!25}.015 & \cellcolor{green!25}.015 & \cellcolor{green!25}.037 \\
                      & Qwen & .595 & \cellcolor{green!25}.014 & \cellcolor{green!25}.046 & \cellcolor{green!25}.049 & \cellcolor{orange!20}.065 & \cellcolor{green!25}.044 & \cellcolor{orange!20}.055 & \cellcolor{green!25}.043 & \cellcolor{orange!20}.053 & .116 & .114 & .134 & .150 \\
                      & Llama & .532 & \cellcolor{green!25}.016 & \cellcolor{green!25}.002 & \cellcolor{green!25}.002 & \cellcolor{green!25}.018 & \cellcolor{green!25}.002 & \cellcolor{green!25}.019 & \cellcolor{green!25}.002 & \cellcolor{green!25}.017 & \cellcolor{green!25}.010 & \cellcolor{green!25}.022 & \cellcolor{green!25}.041 & \cellcolor{orange!20}.061 \\
    \toprule
    \rowcolor{lightgray}\multicolumn{3}{l|}{\textbf{\#settings@5\% significance}} & 12 & 12 & 11 & 13 & 10 & 11 & 12 & 11 & 14 & 12 & \textbf{17} & \textbf{17}\\
    \multicolumn{3}{l|}{\#settings@10\% significance} & 13 & 14 & 16 & 16 & 13 & 13 & 15 & 16 & 17 & 12 & 18 & \textbf{20}\\
    \bottomrule
    \end{tabular}
    \caption{Statistical significance results (p-values) for various Dataset-LLM settings using \textbf{Student's t-test}. The settings with significance at 5\% \& 10\% are shown in \colorbox{green!25}{green} and \colorbox{orange!20}{orange}, respectively. (Penalty functions -- \textbf{B1}: No. of tokens, \textbf{B2}: No. of natural words that are split (baselines); \textbf{AS}/\textbf{UT}/\textbf{PD}: Penalties based based on token anomaly scores, distance from unused tokens, and pairwise distance between tokens, respectively; \textbf{CP}: Contextual penalty)}
    \label{tabResults}
\end{table*}
\subsection{NLP Tasks and Datasets}\label{secDatasets}
We consider a variety of NLP tasks 
and choose one representative dataset from each (Table~\ref{tabTasksAndDatasets}). For all the tasks, tokenization penalties are computed for only the input text (and not for the other details in the prompts like instructions and few-shot examples which are common for all the instances). 
\begin{table}[H]\small
    \centering
    \begin{tabular}{p{0.95\columnwidth}}
    \hline
\textbf{Text Classification}: TREC~\citep{voorhees2000building} \\
\textbf{Text Pair Classification}: RTE~\citep{wang2019glue} \\
\textbf{Passage-based QA}: boolq~\citep{clark2019boolq} \\
\textbf{Factual QA}: Geography~\citep{ramrakhiyani2025gauging} \\
\textbf{NER}: CoNLL 2003~\citep{sang2003introduction} \\
\textbf{Math Reasoning}: GSM~\cite{cobbe2021gsm8k} \\
\textbf{Code Generation}: MBPP~\citep{austin2021program} \\
    \hline
    \end{tabular}
    \caption{NLP Tasks and corresponding datasets (more implementation details in Appendix~\ref{secDatasetDetails} and~\ref{secImplementationDetails})}
    \label{tabTasksAndDatasets}
\end{table}

\subsection{Experimental Results}
\noindent\textbf{Baselines}: We consider two simple baselines for tokenization penalty functions -- (i) \textbf{B1}: computes total number of tokens in the input, and (ii) \textbf{B2}: computes total number of natural words in the input that are split into multiple tokens.\\

\noindent\textbf{Analysis of Results}: Table~\ref{tabResults} reports the results of the Student's t-test\footnote{See Appendix Table~\ref{tabResultsMW} for Mann-Whitney U Test.} for multiple Dataset-LLM combinations for the proposed tokenization penalty functions with various aggregation techniques. For the non-contextual functions, we report only \textit{sum} and \textit{max} due to space constraints. The contextual penalty (CP) function is observed to be the best as it achieves statistical significance at $5\%$ for 17 out of 28 Dataset-LLM combinations. Also, there is agreement between both the statistical tests for most of the combinations. CP is better than other non-contextual functions (AS, UT, and PD) which are better than the baselines. We also carried out ablation study for the design decision of POS importance weight ($wt_p$) used in CP (Table~\ref{tabAblation}). 
Except for RTE and GSM, on all other datasets, CP shows statistically significant effect on the accuracy for most LLMs considered. 
We leave further investigation for RTE and GSM (and necessary changes to the tokenization penalty functions) as a future work.
Figure~\ref{figExampleDecile} shows another interesting analysis where we observe significant difference between accuracy among the top and bottom deciles (most and least challenging instances w.r.t. tokenization) as per CP with \textit{top\_3} aggregation. This difference is observed across all datasets, even for RTE and GSM. 
We also observe that the smaller the vocabulary size of an LLM (i.e. higher its tokenizer fertility~\cite{ali2024tokenizer}), the greater the number of datasets on which tokenization significantly affects 
the LLM's performance, as shown in Figure~\ref{figLLMsVocab}.

\noindent\textbf{Transformation Strategies}: The primary objective of this paper was to establish the statistical significance of the impact of ``bad'' tokenization on model performance. However, it is equally important to explore input transformation strategies to mitigate the issue of bad tokenization. As part of our future work, we plan to investigate several such strategies (illustrated here using examples from Mistral-7B-Instruct\_v0.3):
\begin{itemize}[leftmargin=*, itemsep=0pt]
    \item \textbf{Case modification}. For example, replacing {\it hollywood} (tokenized as {\it hol}, {\it ly}, {\it wood}) with {\it Hollywood} which is a single token.
    \item \textbf{Synonym substitution}. For example, {\it unexceptional} (split as {\it une}, {\it x}, {\it ception}, {\it al}) could be replaced with {\it ordinary} which is single token.
    \item \textbf{Morphology-aware tokenization}. For example, {\it genders} is split as {\it g} + {\it enders}; but a more linguistically informed tokenization would be {\it gender} + {\it s}, recognizing {\it s} as a common plural suffix.
\end{itemize}

\section{Conclusions and Future Work}
Due to the limited vocabulary size of LLMs, natural words are often split into multiple tokens. In this paper, we hypothesized that such splitting of natural words may adversely affect the performance of LLMs on various NLP tasks. To investigate this, we proposed a set of tokenization penalty functions which quantify how ``bad'' is the tokenization for a given text with respect to a specific LLM. We established the statistical significance of our hypothesis on seven different NLP tasks across four different LLMs. 
In future, we plan to explore a few simple 
transformations of input text (as discussed above) that reduce the tokenization penalty and potentially improve model performance.  
Additionally, we believe that insights from this analysis -- both on tokenization issues and mitigation strategies -- can help the design of more effective tokenizers for future LLMs.

\section*{Limitations}
\begin{itemize}[leftmargin=*, itemsep=0pt]
    \item \textbf{Tokenization not being the only issue}: Bad tokenization is not the only cause behind the unexpected performance of LLMs. It is just one of the many causes and that too a weak cause (e.g., there are errors even in instances where there is no tokenization penalty, as seen in Table~\ref{figExamplePerformance}). Hence, attributing some unexpected response by an LLM specifically to only bad tokenization issue is very challenging. What we have attempted to do in this work is only to establish a correlation between tokenization and performance of an LLM on multiple NLP tasks.
    \item \textbf{Number of models and tasks}: In this paper, we have experimented with only a limited number of different LLMs (4) and also a limited number of NLP tasks (7). Although, we have chosen 7 tasks of very different nature and considered 1 representative dataset for each task, there is still some scope of extending the experiments to cover more LLMs, more tasks, and more datasets per task.
    \item \textbf{Closed-source models}: We have not explored closed source models like OpenAI GPT because all our penalty functions need access to embedding matrix (for AS, UT, PD) as well as next token probability assigned to a certain token given its left context (for CP). 
    \item \textbf{Combination of multiple penalty functions}: We have not yet explored how multiple penalty functions can be combined to produce a better combined penalty function. We plan to take this up as a future work.
\end{itemize}


\bibliography{custom}

\appendix

\section{NLP Tasks and Datasets}\label{secDatasetDetails}
In this section, we provide a more detailed description of the NLP tasks and the corresponding datasets used in our experiments.
\begin{itemize}[leftmargin=*, itemsep=0pt]
\item\textbf{Text Classification}: The task is to assign a suitable label to a given input text. We use the test partition of TREC~\cite{voorhees2000building} dataset (\#instances=500).

\item\textbf{Text Pair Classification}: The task is to assign a suitable label to a given pair of two texts. We use the train partition of RTE~\cite{wang2019glue} dataset (\#instances=2490) where for a given pair of premise and hypothesis, the task is to identify whether the premise entails the hypothesis or not. 

\item\textbf{Passage-based QA}: The task is to answer a question based on a specific passage of information. We use the validation partition of boolq dataset~\cite{clark2019boolq} (\#instances=3200).

\item\textbf{Factual QA}: In this task, a factual question is to be answered without any reference passage and only based on the pre-training knowledge of an LLM. We use Geography dataset proposed by~\citet{ramrakhiyani2025gauging} (\#instances=2879).

\item\textbf{Named Entity Recognition (NER)}: The task is to identify named entities in a given text. We use the test partition of CoNLL 2003~\citep{sang2003introduction} dataset (\#instances=1525). We only consider 3 entity types (PER, ORG, and LOC) and discard the sentences having no verb or less than 5 words to retain only meaningful sentences. Unlike all the other tasks where we use an LLM in zero-shot manner, for this NER task we use few-shot examples in TANL format~\cite{paolini2021structured}. The LLM output is considered to be {\em correct} if and only if all the entities in the input are identified correctly (both mention boundaries as well as types).

\item\textbf{Math Reasoning}: The task is to solve a mathematical reasoning question using a multi-step approach. We use the test partition of the Grade School Math (GSM) dataset~\citep{cobbe2021gsm8k} (\#instances=1319). We consider the solution to be {\em correct} if the final answer matches with that of the gold-standard answer, irrespective of the intermediate steps.

\item\textbf{Code Generation}: The task to generate suitable python code for the given problem. We used the sanitized (hand-verified) version of the MBPP (Mostly Basic Python Problems) dataset~\citep{austin2021program} (\#instances=427). Each instance in the dataset is also accompanied by 3 test cases. We consider the generated code to be {\em correct} if it passes these 3 test cases.
\end{itemize}

\section{Implementation Details}\label{secImplementationDetails}
For obtaining natural words, we use spaCy tokenizer from the model {\small\tt en\_core\_web\_trf-3.8.0}.

We use the following 4 models for all our experiments having 8 billion or less parameters because of limited hardware available with us.
\begin{itemize}[leftmargin=*, itemsep=0pt]
    \item \textbf{Phi}: Phi-3.5-mini-instruct\footnote{\url{https://huggingface.co/microsoft/Phi-3.5-mini-instruct}} (3.8 billion parameters)
    \item \textbf{Mistral}: Mistral-7B-Instruct-v0.3\footnote{\url{https://huggingface.co/mistralai/Mistral-7B-Instruct-v0.3}} (7 billion parameters)
    \item \textbf{Qwen}: Qwen2.5-7B-Instruct\footnote{\url{https://huggingface.co/Qwen/Qwen2.5-7B-Instruct}} (7 billion parameters)
    \item \textbf{Llama}: Meta-Llama-3-8B-Instruct\footnote{\url{https://huggingface.co/meta-llama/Meta-Llama-3-8B-Instruct}} (8 billion parameters)
\end{itemize}
\noindent\textbf{Hyperparameters}: We only used these LLMs in inference mode and uniformly used the temperature setting of 0 for all the 4 LLMs across all the 7 tasks to ensure more consistent and repeatable responses (based on the recommendation by~\citet{renze2024effect}).\\

\noindent\textbf{GPU size}: All the experiments were run on an Nvidia Tesla V100 with 32GB GPU RAM.\\

\noindent\textbf{Statistical tests}: We used implementations of the Student's t-test and Mann Whitney U test available as part of the stats package of scipy\footnote{\url{https://docs.scipy.org/doc/scipy/reference/generated/scipy.stats.ttest_ind.html}} \footnote{\url{https://docs.scipy.org/doc/scipy/reference/generated/scipy.stats.mannwhitneyu.html}}.

\section{Prompts used for the NLP tasks}
Table~\ref{tabLLMPrompts} shows the detailed prompts used for each NLP task in our experiments. Zero-shot prompting is used for all the tasks except for NER where we use 8 few-shot examples.

\begin{table*}[t]\small
    \centering
    \begin{tabular}{p{0.95\linewidth}}
    \toprule
    \textbf{boolq dataset}:\\
    Answer the Question below as Yes or No, based on the following Passage. \\
Question: {\color{blue}\{question\}} \\
Passage: {\color{blue}\{passage\}} \\
Answer: \\
    \midrule
    \textbf{Geography dataset}:\\
    Answer for the following question in one word.  {\color{blue}\{question\}} \\
Answer:\\
    \midrule
    \textbf{RTE dataset}:\\
    The Entailment relation holds between a Premise and a Hypothesis when the Hypothesis can be inferred to be true if the Premise is true. For the following pair of PREMISE and HYPOTHESIS, identify whether Entailment relation holds or not. Do not generate any extra text. Only answer as "Yes" (if Entailment holds) or "No" (if Entailment does not hold).\\
PREMISE: {\color{blue}\{premise\}}\\
HYPOTHESIS: {\color{blue}\{hypothesis\}} \\
Entailment:\\
    
    \midrule
    \textbf{TREC dataset}:\\
    There are 6 different possible answer types for any Question as follows:\\
1. ABBR: an abbreviation\\
2. ENTY: an entity\\
3. DESC: a description or abstract concept\\
4. HUM: a human being\\
5. LOC: a geographical location\\
6. NUM: a numeric value\\
What is the suitable answer type from the above list for the following Question? \\
Question: {\color{blue}\{question\}}\\
Answer type:\\
    \midrule
    \textbf{CoNLL dataset}:\\
    You are given a SENTENCE, your task is to identify the entites ['person', 'organisation', 'location'] present in the SENTENCE. Avoid mentioning any other entites in the response. Only identify the entites provided in the list. \\
Some examples of your task are given below-\\
\textbf{SENTENCE}: Enn Markvart , chairman of the National Election Commission said 96 members of parliament cast votes .\\
\textbf{OUTPUT}: [ Enn Markvart | person ] , chairman of the [ National Election Commission | organization ] said 96 members of parliament cast votes .\\
\textbf{SENTENCE}: He did not elaborate . \\
\textbf{OUTPUT}: He did not elaborate . \\
$\cdots$ \textit{\color{blue}\{6 more few-shot examples\}} \\






Now, identify the entities from-\\
SENTENCE: {\color{blue}\{sentence\}}\\
    \midrule
\textbf{GSM dataset}:\\
Solve the following mathematical Question by using step-by-step reasoning. Write the final answer following the prefix "Final Answer:" \\
Question: {\color{blue}\{question\}} \\
    \midrule
\textbf{MBPP dataset}:\\
{\color{blue}\{problem\}} \\
Use the following function name: {\color{blue}\{function\_name\}} \\
Generate only executable python code and nothing else. \\
    \bottomrule
    \end{tabular}
    \caption{Prompts used for various NLP tasks}
    \label{tabLLMPrompts}
\end{table*}

\section{Examples of Penalty Functions}
Table~\ref{tabExamplePenaltyFunctions} shows 2 example sentences from the Geography dataset. For each example, it shows how each LLM tokenizes the input sentence (in terms of natural words that are split and also the values for each proposed penalty function.

\begin{table*}\small\center
\begin{tabular}{p{0.9\linewidth}}
\toprule
\rowcolor{lightgray}\textbf{Sentence 1}: \textit{In which country, is the Raysko Praskalo waterfall located?} \\
\rowcolor{lightgray}\textbf{Gold answer}: \textit{Bulgaria} \\
\midrule
\textbf{LLM}: Phi, \textbf{Generated answer}: \textit{Slovenia}, 
\textbf{Split natural words}: \textit{\color{blue}\_Ray}, \textit{\color{blue}sko}, \textit{\color{blue}\_Pr}, \textit{\color{blue}ask}, \textit{\color{blue}alo}, \textit{\color{blue}\_water}, \textit{\color{blue}fall} \\
\textbf{Tokenization Penalties}: AS: 0.644, UT: 0.405, PD: 0.978, CP: 1.72\\
\midrule
\textbf{LLM}: Mistral, \textbf{Generated answer}: \textit{Russia}, 
\textbf{Split natural words}: \textit{\color{blue}\_R}, \textit{\color{blue}ays}, \textit{\color{blue}ko}, \textit{\color{blue}\_Pr}, \textit{\color{blue}ask}, \textit{\color{blue}alo}, \textit{\color{blue}\_water}, \textit{\color{blue}fall} \\
\textbf{Tokenization Penalties}: AS: 0.342, UT: 0.466, PD: 1.022, CP: 1.87\\
\midrule
\textbf{LLM}: Qwen, \textbf{Generated answer}: \textit{Bosnia and Herzegovina}, 
\textbf{Split natural words}: \textit{\color{blue}\_Rays}, \textit{\color{blue}ko}, \textit{\color{blue}\_Pr}, \textit{\color{blue}ask}, \textit{\color{blue}alo}\\
\textbf{Tokenization Penalties}: AS: 0.387, UT: 0.527, PD: 0.982, CP: 0.9999\\
\midrule
\textbf{LLM}: Llama, \textbf{Generated answer}: \textit{Serbia}, 
\textbf{Split natural words}: \textit{\color{blue}\_Rays}, \textit{\color{blue}ko}, \textit{\color{blue}\_Pr}, \textit{\color{blue}ask}, \textit{\color{blue}alo}\\
\textbf{Tokenization Penalties}: AS: 0.55, UT: 0.47, PD: 1.00, CP: 0.712
\\
\midrule
\midrule
\rowcolor{lightgray}\textbf{Sentence 2}: \textit{In which country, is the city of Helsinki located?} \\
\rowcolor{lightgray}\textbf{Gold answer}: \textit{Finland} \\
\midrule
\textbf{LLM}: Phi, \textbf{Generated answer}: \textit{Finland}, 
\textbf{Split natural words}: \textit{\color{blue}\_Hels}, \textit{\color{blue}ink}, \textit{\color{blue}i}\\
\textbf{Tokenization Penalties}: AS: 0.597, UT: 0.341, PD: 0.957, CP: 0.404\\
\midrule
\textbf{LLM}: Mistral, \textbf{Generated answer}: \textit{Finland}, 
\textbf{Split natural words}: \textit{\color{blue}\_H}, \textit{\color{blue}els}, \textit{\color{blue}ink}, \textit{\color{blue}i} \\
\textbf{Tokenization Penalties}: AS: 0.308, UT: 0.362, PD: 0.994, CP: 0.587\\
\midrule
\textbf{LLM}: Qwen, \textbf{Generated answer}: \textit{Finland}, 
\textbf{Split natural words}: NONE \\
\textbf{Tokenization Penalties}: AS: 0, UT: 0, PD: 0, CP: 0\\
\midrule
\textbf{LLM}: Llama, \textbf{Generated answer}: \textit{Finland}, 
\textbf{Split natural words}: NONE \\
\textbf{Tokenization Penalties}: AS: 0, UT: 0, PD: 0, CP: 0\\
\bottomrule
\end{tabular}
\caption{Example sentences from the Geography and their tokenization penalties using max aggregation. All 4 LLMs generate incorrect response for Sentence 1 whereas all the 4 LLMs generated a correct response for Sentence 2. Some key observations: (i) The penalty scores are NOT comparable across the penalty functions and also across the LLMs. (ii) The penalty scores for a specific penalty function and a specific LLM are comparable across multiple sentences (inputs). In this example, it can be observed that penalty for Sentence 2 is consistently lower than the penalty for Sentence 1 for each penalty function and for each LLM. (iii) As Qwen has much larger vocabulary size as compared to Phi and Mistral, it tends to produce lesser number of tokens.}
\label{tabExamplePenaltyFunctions}
\end{table*}

\begin{table*}[t]\small
    \centering
    \begin{tabular}{p{1cm}lc|c|c|cc|cc|cc|cccc}
    \toprule
    \multirow{2}{*}{\textbf{Dataset}} & \multirow{2}{*}{\textbf{Model}} & \multirow{2}{*}{\textbf{Acc}} & \multirow{2}{*}{\textbf{B1}} & \multirow{2}{*}{\textbf{B2}} & \multicolumn{2}{c|}{\textbf{AS}} & \multicolumn{2}{c|}{\textbf{UT}} & \multicolumn{2}{c|}{\textbf{PD}} & \multicolumn{4}{c}{\textbf{CP}} \\
    \cline{6-15}
     &  &  &  &  & \textbf{sum} & \textbf{max} & \textbf{sum} & \textbf{max} & \textbf{sum} & \textbf{max} & \textbf{sum} & \textbf{avg} & \textbf{max} & \textbf{top3} \\
    \midrule
    \multirow{4}{*}{TREC} & Phi & .796 & \cellcolor{green!25}.001 & \cellcolor{orange!20}.082 & \cellcolor{orange!20}.058 & .160 & .185 & .260 & \cellcolor{green!25}.033 & \cellcolor{green!25}.019 & \cellcolor{green!25}.027 & .344 & \cellcolor{green!25}.034 & \cellcolor{green!25}.038 \\
                      & Mistral & .710 & .290 & \cellcolor{orange!20}.084 & \cellcolor{green!25}.004 & \cellcolor{green!25}.001 & .141 & \cellcolor{orange!20}.079 & \cellcolor{orange!20}.087 & \cellcolor{orange!20}.082 & \cellcolor{green!25}.037 & \cellcolor{green!25}.025 & \cellcolor{green!25}.039 & \cellcolor{green!25}.029 \\
                      & Qwen & .776 & .318 & \cellcolor{orange!20}.081 & \cellcolor{orange!20}.087 & \cellcolor{orange!20}.096 & .102 & \cellcolor{orange!20}.095 & \cellcolor{orange!20}.088 & \cellcolor{orange!20}.082 & \cellcolor{orange!20}.059 & \cellcolor{green!25}.048 & \cellcolor{orange!20}.061 & \cellcolor{orange!20}.058 \\
                      & Llama & .754 & .229 & .161 & .129 & .122 & .200 & .186 & .129 & .120 & .129 & \cellcolor{orange!20}.069 & .127 & .119 \\
    \midrule
    \multirow{4}{*}{RTE} & Phi & .865 & .328 & .389 & .433 & .871 & .383 & .161 & .412 & .507 & .184 & .164 & .321 & .282 \\
                      & Mistral & .818 & .874 & .976 & .978 & .922 & .955 & .346 & .982 & .530 & .833 & .659 & .625 & .409 \\
                      & Qwen & .895 & .584 & .906 & .914 & .486 & .915 & .778 & .932 & .924 & .754 & .587 & .750 & .536 \\
                      & Llama & .748 & .999 & .999 & .999 & .984 & .999 & .985 & .999 & .999 & .962 & .266 & .567 & .390 \\
    \midrule
    \multirow{4}{*}{boolq} & Phi & .757 & .169 & .166 & .164 & \cellcolor{green!25}.019 & .238 & .966 & .151 & .366 & \cellcolor{green!25}.002 & \cellcolor{green!25}.000 & \cellcolor{green!25}.000 & \cellcolor{green!25}.000 \\
                      & Mistral & .755 & .296 & .168 & .149 & \cellcolor{green!25}.038 & .193 & .742 & .151 & .149 & \cellcolor{green!25}.026 & \cellcolor{green!25}.006 & \cellcolor{orange!20}.065 & \cellcolor{green!25}.015 \\
                      & Qwen & .773 & .695 & .606 & .594 & .529 & .642 & .946 & .618 & .971 & \cellcolor{orange!20}.077 & \cellcolor{green!25}.048 & \cellcolor{green!25}.005 & \cellcolor{green!25}.009 \\
                      & Llama & .798 & .658 & .540 & .470 & .226 & .608 & .949 & .562 & .897 & \cellcolor{orange!20}.090 & \cellcolor{orange!20}.073 & \cellcolor{green!25}.012 & \cellcolor{green!25}.008 \\
    \midrule
    \multirow{4}{*}{\footnotesize Geography} & Phi & .751 & .400 & \cellcolor{green!25}.002 & \cellcolor{green!25}.020 & .289 & .100 & .671 & \cellcolor{green!25}.006 & \cellcolor{orange!20}.061 & \cellcolor{green!25}.000 & \cellcolor{green!25}.000 & \cellcolor{green!25}.000 & \cellcolor{green!25}.000 \\
                      & Mistral & .790 & .783 & \cellcolor{green!25}.000 & \cellcolor{green!25}.001 & .175 & \cellcolor{green!25}.029 & .725 & \cellcolor{green!25}.000 & \cellcolor{green!25}.000 & \cellcolor{green!25}.000 & \cellcolor{green!25}.000 & \cellcolor{green!25}.000 & \cellcolor{green!25}.000 \\
                      & Qwen & .726 & .930 & \cellcolor{green!25}.000 & \cellcolor{green!25}.000 & \cellcolor{green!25}.000 & \cellcolor{green!25}.006 & \cellcolor{green!25}.002 & \cellcolor{green!25}.000 & \cellcolor{green!25}.000 & \cellcolor{green!25}.000 & \cellcolor{green!25}.000 & \cellcolor{green!25}.000 & \cellcolor{green!25}.000 \\
                      & Llama & .699 & .978 & .992 & .997 & .942 & .977 & .778 & .857 & .365 & .425 & .664 & \cellcolor{green!25}.025 & \cellcolor{green!25}.002 \\
    \midrule
    \multirow{4}{*}{CoNLL} & Phi & .280 & \cellcolor{green!25}.000 & \cellcolor{green!25}.000 & \cellcolor{green!25}.000 & \cellcolor{green!25}.003 & \cellcolor{green!25}.000 & \cellcolor{green!25}.000 & \cellcolor{green!25}.000 & \cellcolor{orange!20}.053 & \cellcolor{green!25}.000 & .363 & \cellcolor{green!25}.000 & \cellcolor{green!25}.013 \\
                      & Mistral & .336 & \cellcolor{green!25}.000 & \cellcolor{green!25}.000 & \cellcolor{green!25}.000 & \cellcolor{green!25}.000 & \cellcolor{green!25}.000 & \cellcolor{green!25}.000 & \cellcolor{green!25}.000 & \cellcolor{green!25}.000 & \cellcolor{green!25}.000 & .424 & \cellcolor{green!25}.000 & \cellcolor{green!25}.000 \\
                      & Qwen & .481 & \cellcolor{green!25}.000 & \cellcolor{green!25}.000 & \cellcolor{green!25}.000 & \cellcolor{green!25}.000 & \cellcolor{green!25}.000 & \cellcolor{green!25}.000 & \cellcolor{green!25}.000 & \cellcolor{green!25}.000 & \cellcolor{green!25}.000 & .067 & \cellcolor{green!25}.000 & \cellcolor{green!25}.013 \\
                      & Llama & .389 & \cellcolor{green!25}.000 & \cellcolor{green!25}.000 & \cellcolor{green!25}.000 & \cellcolor{green!25}.000 & \cellcolor{green!25}.000 & \cellcolor{green!25}.000 & \cellcolor{green!25}.000 & \cellcolor{green!25}.000 & \cellcolor{green!25}.000 & .095 & \cellcolor{green!25}.003 & \cellcolor{green!25}.014 \\
    \midrule
    \multirow{4}{*}{GSM} & Phi & .867 & \cellcolor{green!25}.000 & \cellcolor{orange!20}.075 & \cellcolor{orange!20}.080 & .509 & \cellcolor{orange!20}.066 & .192 & \cellcolor{orange!20}.076 & \cellcolor{orange!20}.089 & \cellcolor{green!25}.004 & .677 & \cellcolor{orange!20}.060 & \cellcolor{green!25}.011 \\
                      & Mistral & .640 & \cellcolor{green!25}.000 & .117 & .181 & .848 & .147 & .554 & \cellcolor{orange!20}.097 & .274 & .201 & .996 & .539 & .269 \\
                      & Qwen & .916 & \cellcolor{green!25}.000 & .625 & .703 & .723 & .604 & .378 & .682 & .635 & .427 & .830 & .352 & .247 \\
                      & Llama & .813 & \cellcolor{green!25}.000 & .205 & .249 & .185 & .246 & .295 & .164 & \cellcolor{orange!20}.070 & .121 & .771 & .137 & \cellcolor{orange!20}.097 \\
    \midrule
    \multirow{4}{*}{MBPP} & Phi & .553 & .270 & \cellcolor{green!25}.000 & \cellcolor{green!25}.000 & \cellcolor{green!25}.000 & \cellcolor{green!25}.000 & \cellcolor{green!25}.000 & \cellcolor{green!25}.000 & \cellcolor{green!25}.000 & \cellcolor{green!25}.000 & \cellcolor{green!25}.000 & \cellcolor{green!25}.000 & \cellcolor{green!25}.001 \\
                      & Mistral & .372 & \cellcolor{green!25}.007 & \cellcolor{green!25}.000 & \cellcolor{green!25}.000 & \cellcolor{green!25}.003 & \cellcolor{green!25}.000 & \cellcolor{green!25}.000 & \cellcolor{green!25}.000 & \cellcolor{green!25}.003 & \cellcolor{green!25}.002 & \cellcolor{green!25}.006 & \cellcolor{green!25}.021 & \cellcolor{green!25}.046 \\
                      & Qwen & .595 & .117 & \cellcolor{orange!20}.054 & \cellcolor{orange!20}.055 & \cellcolor{orange!20}.064 & \cellcolor{orange!20}.060 & \cellcolor{orange!20}.057 & \cellcolor{green!25}.039 & \cellcolor{green!25}.043 & \cellcolor{orange!20}.056 & \cellcolor{orange!20}.073 & \cellcolor{orange!20}.058 & \cellcolor{orange!20}.061 \\
                      & Llama & .532 & \cellcolor{orange!20}.062 & \cellcolor{green!25}.014 & \cellcolor{green!25}.013 & \cellcolor{green!25}.019 & \cellcolor{green!25}.014 & \cellcolor{green!25}.023 & \cellcolor{green!25}.009 & \cellcolor{green!25}.011 & \cellcolor{green!25}.018 & \cellcolor{green!25}.019 & \cellcolor{green!25}.026 & \cellcolor{green!25}.031 \\
    \toprule
    \rowcolor{lightgray}\multicolumn{3}{l|}{\textbf{\#settings@5\% significance}} & 10 & 10 & 11 & 11 & 9 & 8 & 12 & 10 & 15 & 11 & 16 & \textbf{18} \\
    \multicolumn{3}{l|}{\#settings@10\% significance} & 11 & 15 & 15 & 13 & 12 & 11 & 16 & 16 & 19 & 16 & 20 & \textbf{21} \\
    \bottomrule
    \end{tabular}
    \caption{Statistical significance results (p-values) for various Dataset-LLM settings using \textbf{Mann-Whitney U Test}. The settings with significance at 5\% \& 10\% are shown in \colorbox{green!25}{green} and \colorbox{orange!20}{orange}, respectively. (Penalty functions -- B1: Baseline of no. of tokens, B2: Baseline of no. of natural words that are split, AS: Penalty based on token anomaly scores, UT: Penalty based on distance from unused tokens, PD: Penalty based on pairwise distance between tokens, CP: Contextual penalty)}
    \label{tabResultsMW}
\end{table*}

\begin{table*}[t]\small
    \centering
    \begin{tabular}{p{1.1cm}lc|cccc|cccc}
    \toprule
    \multirow{2}{*}{\textbf{Dataset}} & \multirow{2}{*}{\textbf{Model}} & \multirow{2}{*}{\textbf{Acc}} & \multicolumn{4}{c|}{\textbf{CP} (with POS multiplier)} & \multicolumn{4}{c}{\textbf{CP} (without POS multiplier)} \\
    \cline{4-11}
     &  &  & \textbf{sum} & \textbf{avg} & \textbf{max} & \textbf{top3} & \textbf{sum} & \textbf{avg} & \textbf{max} & \textbf{top3}\\
    \midrule
    \multirow{4}{*}{TREC} & Phi & .796 & \cellcolor{orange!20}.068 & .746 & \cellcolor{green!25}.040 & \cellcolor{green!25}.043 & \cellcolor{orange!20}.060 & .593 & \cellcolor{green!25}.037 & \cellcolor{orange!20}.065 \\
                      & Mistral & .710 & .110 & \cellcolor{green!25}.040 & \cellcolor{green!25}.033 & \cellcolor{green!25}.028 & \cellcolor{orange!20}.083 & \cellcolor{green!25}.020 & \cellcolor{green!25}.026 & \cellcolor{green!25}.027 \\
                      & Qwen & .776 & \cellcolor{green!25}.037 & \cellcolor{green!25}.019 & \cellcolor{green!25}.039 & \cellcolor{green!25}.033 & \cellcolor{green!25}.024 & \cellcolor{green!25}.009 & \cellcolor{green!25}.025 & \cellcolor{green!25}.021 \\
                      & Llama & .754 & \cellcolor{orange!20}.079 & \cellcolor{green!25}.026 & \cellcolor{orange!20}.076 & \cellcolor{orange!20}.069 & .109 & \cellcolor{green!25}.019 & .101 & \cellcolor{orange!20}.094 \\
    \midrule
    \multirow{4}{*}{RTE} & Phi & .865 & .109 & .150 & .263 & .242 & .102 & .140 & .266 & .290 \\
                      & Mistral & .818 & .830 & .632 & .560 & .459 & .830 & .644 & .428 & .389 \\
                      & Qwen & .895 & .781 & .542 & .670 & .574 & .762 & .503 & .585 & .552 \\
                      & Llama & .748 & .971 & .291 & .639 & .469 & .995 & .462 & .899 & .774 \\
    \midrule
    \multirow{4}{*}{boolq} & Phi & .757 & \cellcolor{green!25}.011 & \cellcolor{green!25}.001 & \cellcolor{green!25}.000 & \cellcolor{green!25}.000 & \cellcolor{green!25}.008 & \cellcolor{green!25}.000 & \cellcolor{green!25}.003 & \cellcolor{green!25}.000 \\
                      & Mistral & .755 & \cellcolor{green!25}.045 & \cellcolor{green!25}.006 & \cellcolor{green!25}.015 & \cellcolor{green!25}.012 & \cellcolor{green!25}.035 & \cellcolor{green!25}.003 & \cellcolor{green!25}.011 & \cellcolor{green!25}.008 \\
                      & Qwen & .773 & \cellcolor{orange!20}.059 & \cellcolor{green!25}.037 & \cellcolor{green!25}.009 & \cellcolor{green!25}.018 & .137 & \cellcolor{orange!20}.070 & \cellcolor{orange!20}.084 & .178 \\
                      & Llama & .798 & .107 & .122 & \cellcolor{green!25}.018 & \cellcolor{green!25}.012 & \cellcolor{orange!20}.050 & \cellcolor{orange!20}.062 & .114 & \cellcolor{orange!20}.057 \\
    \midrule
    \multirow{4}{*}{\footnotesize Geography} & Phi & .751 & \cellcolor{green!25}.000 & \cellcolor{green!25}.000 & \cellcolor{green!25}.000 & \cellcolor{green!25}.000 & \cellcolor{green!25}.000 & \cellcolor{green!25}.000 & \cellcolor{green!25}.000 & \cellcolor{green!25}.000 \\
                      & Mistral & .790 & \cellcolor{green!25}.000 & \cellcolor{green!25}.000 & \cellcolor{green!25}.000 & \cellcolor{green!25}.000 & \cellcolor{green!25}.000 & \cellcolor{green!25}.000 & \cellcolor{green!25}.000 & \cellcolor{green!25}.000 \\
                      & Qwen & .726 & \cellcolor{green!25}.000 & \cellcolor{green!25}.000 & \cellcolor{green!25}.000 & \cellcolor{green!25}.000 & \cellcolor{green!25}.000 & \cellcolor{green!25}.000 & \cellcolor{green!25}.000 & \cellcolor{green!25}.000 \\
                      & Llama & .699 & .756 & .645 & .207 & \cellcolor{orange!20}.078 & .734 & .607 & .186 & \cellcolor{orange!20}.077 \\
    \midrule
    \multirow{4}{*}{CoNLL} & Phi & .280 & \cellcolor{green!25}.000 & .336 & \cellcolor{green!25}.001 & \cellcolor{green!25}.012 & \cellcolor{green!25}.000 & .454 & .265 & .464 \\
                      & Mistral & .336 & \cellcolor{green!25}.000 & .358 & \cellcolor{green!25}.000 & \cellcolor{green!25}.000 & \cellcolor{green!25}.000 & .515 & \cellcolor{green!25}.000 & \cellcolor{green!25}.000 \\
                      & Qwen & .481 & \cellcolor{green!25}.000 & .114 & \cellcolor{green!25}.000 & \cellcolor{green!25}.030 & \cellcolor{green!25}.000 & .116 & \cellcolor{green!25}.012 & .158 \\
                      & Llama & .389 & \cellcolor{green!25}.000 & .312 & \cellcolor{green!25}.000 & \cellcolor{green!25}.011 & \cellcolor{green!25}.000 & .243 & \cellcolor{green!25}.000 & \cellcolor{green!25}.005 \\
    \midrule
    \multirow{4}{*}{GSM} & Phi & .867 & \cellcolor{green!25}.009 & .666 & .205 & \cellcolor{green!25}.040 & \cellcolor{green!25}.006 & .590 & .211 & \cellcolor{orange!20}.056 \\
                      & Mistral & .640 & .119 & .998 & .434 & .265 & \cellcolor{green!25}.039 & .991 & .191 & \cellcolor{orange!20}.095 \\
                      & Qwen & .916 & .426 & .929 & .493 & .230 & .491 & .937 & .504 & .283 \\
                      & Llama & .813 & .134 & .954 & .198 & .116 & \cellcolor{orange!20}.099 & .970 & .122 & \cellcolor{orange!20}.076 \\
    \midrule
    \multirow{4}{*}{MBPP} & Phi & .553 & \cellcolor{green!25}.000 & \cellcolor{green!25}.000 & \cellcolor{green!25}.000 & \cellcolor{green!25}.002 & \cellcolor{green!25}.000 & \cellcolor{green!25}.000 & \cellcolor{green!25}.000 & \cellcolor{green!25}.001 \\
                      & Mistral & .372 & \cellcolor{green!25}.004 & \cellcolor{green!25}.015 & \cellcolor{green!25}.015 & \cellcolor{green!25}.037 & \cellcolor{green!25}.002 & \cellcolor{green!25}.009 & \cellcolor{green!25}.010 & \cellcolor{green!25}.025 \\
                      & Qwen & .595 & .116 & .114 & .134 & .150 & \cellcolor{orange!20}.082 & \cellcolor{orange!20}.080 & \cellcolor{orange!20}.096 & .106 \\
                      & Llama & .532 & \cellcolor{green!25}.010 & \cellcolor{green!25}.022 & \cellcolor{green!25}.041 & \cellcolor{orange!20}.061 & \cellcolor{green!25}.005 & \cellcolor{green!25}.011 & \cellcolor{green!25}.022 & \cellcolor{green!25}.037 \\
    \toprule
    \rowcolor{lightgray}\multicolumn{3}{l|}{\textbf{\#settings@5\% significance}} & 14 & 12 & \textbf{17} & \textbf{17} & 15 & 11 & 14 & 12 \\
    \midrule
    \multicolumn{3}{l|}{\#settings@10\% significance} & 17 & 12 & 18 & \textbf{20} & 20 & 14 & 16 & 19 \\
    \bottomrule
    \end{tabular}
    \caption{Ablation study for using POS importance weight ($wt_p$) in the Contextual Penalty (CP) function. Statistical significance results (p-values) for various Dataset-LLM settings using \textbf{Student's t-test} are shown. The settings with significance at 5\% \& 10\% are shown in \colorbox{green!25}{green} and \colorbox{orange!20}{orange}, respectively.}
    \label{tabAblation}
\end{table*}

\begin{figure}
    \centering
    \includegraphics[width=\columnwidth,height=0.5\columnwidth]{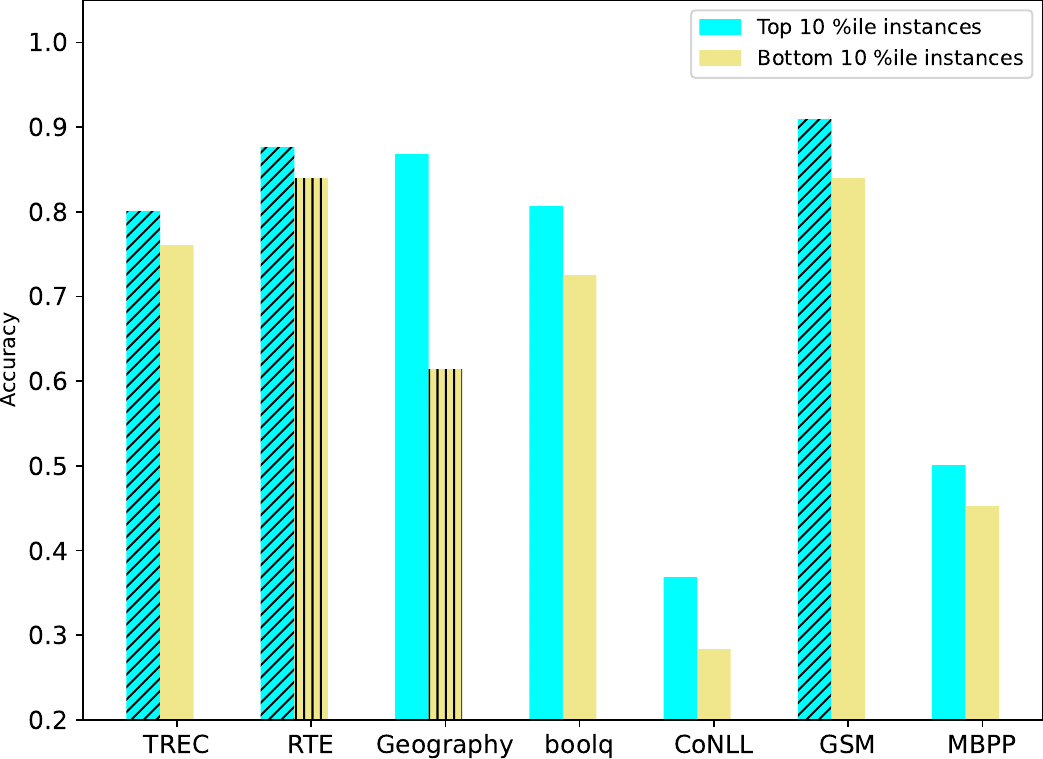}
    \caption{Accuracy difference between top and bottom deciles of the instances as per CP (top\_3) for Phi}
    \label{figExampleDecile}
\end{figure}
\begin{figure}
    \centering
    \includegraphics[width=\columnwidth,height=0.55\columnwidth]{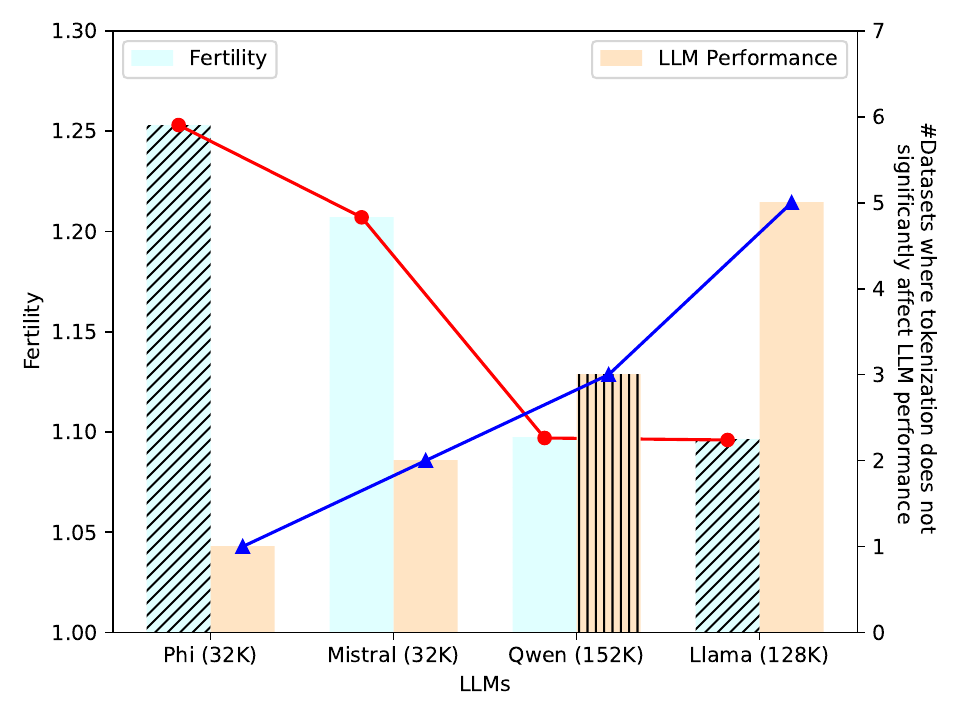}
    \caption{Comparing LLMs across multiple aspects: fertility (average number of tokens per natural word across all the datasets~\cite{ali2024tokenizer}) and performance (number of datasets where the LLM is not significantly affected by tokenization, as per CP (top3) penalty function). The numbers in bracket after each LLM indicate its vocabulary size.}
    \label{figLLMsVocab}
\end{figure}

\section{Comparing Contextual Penalty (CP) with Perplexity}\label{secPPL}
Although, the formula for computing the tokenization penalty $CP$ (Eq.~\ref{eqCP}) looks similar to {\em perplexity}, there are some key differences. Perplexity quantifies how much an LLM is {\em perplexed} (surprised) to observe a certain text whereas $CP$ quantifies {\em goodness} of tokenization for a certain text for a specific LLM. Therefore, $CP$ only considers those {\em natural words} which are split in the tokenization process and computes the penalty for each such word (which is further aggregated using some aggregation function to compute overall penalty for the text). Those natural words which are not split, contribute zero penalty as per $CP$. Moreover, $CP$ also weighs importance of each natural word based on its POS tag. On the other hand, perplexity computation considers {\em all} tokens irrespective of whether a token corresponds to a natural word or just a part of it. It neither considers the concept of natural words nor POS tagging information. Overall, perplexity and $CP$ both try to quantify how well an LLM can comprehend certain text but $CP$ specifically focuses on the tokenization aspect. Similar to the tokenization penalty functions, perplexity may also affect the LLM's performance on various tasks which we show in Table~\ref{tabPPL} using the similar statistical significance test. 

The difference between CP and perplexity can also be seen in the results of the statistical test (Tables~\ref{tabResults} and~\ref{tabPPL}) where CP seems to have a significant effect on the performance for the TREC dataset but perplexity does not. On the other hand, for RTE-Llama combination, perplexity has a significant effect on performance but CP does not.

\begin{table}[bp]\small
    \centering
    \begin{tabular}{p{1cm}lc|c}
    \toprule
    \textbf{Dataset} & \textbf{Model} & \textbf{Acc} & \textbf{Perplexity} \\
    \midrule
    \multirow{4}{*}{TREC} & Phi & .796 & .161 \\
                         & Mistral & .710 & .767 \\
                         & Qwen & .776 & .118 \\
                         & Llama & .754 & .337 \\
    \midrule
    \multirow{4}{*}{RTE} & Phi & .865 & .122 \\
                         & Mistral & .818 & .205 \\
                         & Qwen & .895 & .723 \\
                         & Llama & .748 & \cellcolor{green!25}.001 \\
    \midrule
    \multirow{4}{*}{boolq} & Phi & .757 & \cellcolor{green!25}.000 \\
                         & Mistral & .755 & \cellcolor{green!25}.007  \\
                         & Qwen & .773 & \cellcolor{green!25}.009 \\
                         & Llama & .798 & \cellcolor{green!25}.000 \\
    \midrule
    \multirow{4}{*}{Geography} & Phi & .751 & \cellcolor{green!25}.000 \\
                         & Mistral & .790 & \cellcolor{green!25}.000 \\
                         & Qwen & .726 & \cellcolor{green!25}.000 \\
                         & Llama & .699 & \cellcolor{green!25}.000 \\
    \midrule
    \multirow{4}{*}{CoNLL} & Phi & .280 & .635 \\
                         & Mistral & .336 & .999 \\
                         & Qwen & .481 & .877 \\
                         & Llama & .389 & .994 \\
    \midrule
    \multirow{4}{*}{GSM} & Phi & .867 & \cellcolor{green!25}.000 \\
                         & Mistral & .640 & \cellcolor{green!25}.002 \\
                         & Qwen & .916 & \cellcolor{green!25}.025 \\
                         & Llama & .813 & .257 \\
    \midrule
    \multirow{4}{*}{MBPP} & Phi & .553 & \cellcolor{green!25}.000 \\
                         & Mistral & .372 & .276 \\
                         & Qwen & .595 & \cellcolor{green!25}.000 \\
                         & Llama & .532 & \cellcolor{green!25}.000 \\
    \toprule
    \rowcolor{lightgray}\multicolumn{3}{l|}{\textbf{\#settings@5\% significance}} & 15 \\
    \midrule
    \multicolumn{3}{l|}{\#settings@10\% significance} & 15 \\
    \bottomrule
    \end{tabular}
    \caption{Statistical significance results (p-values) for various Dataset-LLM settings using \textbf{Student's t-test} for Perplexity}
    \label{tabPPL}
\end{table}

\section{Background concepts}
\subsection{Isolation Forest (IF)~\cite{liu2008isolation}}\label{secIF}
It is an anomaly detection technique based on the idea that anomalies are data points or instances that are few and different. It works by randomly selecting a feature and then randomly selecting a split value between the maximum and minimum values of that feature. This process is repeated to build an ensemble of isolation trees. The anomaly score for each instance is computed as follows:
\begin{enumerate}
    \item Each instance is passed through the trees, and the {\em path length} (number of splits required to isolate the instance) is recorded.
    \item {\em Anomalies} tend to have shorter path lengths because they are easier to isolate.
    \item The anomaly score for an instance is calculated based on the average path length across all trees. A higher score (closer to 1) indicates a higher likelihood of that instance being an anomaly.
\end{enumerate}
In our case, each token is represented by a feature vector (as per the output embedding matrix of an LLM). Isolation Forest is then run over all the tokens in this space and anomaly score is computed for each token. The tokens with higher anomaly score are supposed to be more anomalous than other normal tokens with lower anomaly scores.

\subsection{Byte Pair Encoding (BPE)~\cite{sennrich2016neural}}
It is a simple and effective text tokenization algorithm used for building the vocabulary for most LLMs. It compresses text by iteratively replacing the most frequent pair of adjacent symbols (characters or character sequences) with a new symbol. It starts with a sequence of characters, counts all adjacent symbol pairs, and then replaces the most frequent pair with a new token. This process is repeated for a fixed number of steps or until no more pairs are left. This results in a vocabulary of subword units that balances between characters and full words, making it especially useful for handling rare or unknown words in NLP models.

\end{document}